\documentclass[letterpaper]{article} 
\usepackage{aaai25}  
\usepackage{times}  
\usepackage{helvet}  
\usepackage{courier}  
\usepackage[hyphens]{url}  
\usepackage{graphicx} 
\usepackage{natbib}  
\usepackage{caption} 
\usepackage{algorithm}
\usepackage{algorithmic}
\usepackage{booktabs}
\usepackage{varioref}
\usepackage{booktabs}  
\usepackage{array}     
\usepackage{arydshln}  
\usepackage{enumitem}

\usepackage{newfloat}
\usepackage{listings}

\usepackage{subcaption}

\DeclareCaptionStyle{ruled}{labelfont=normalfont,labelsep=colon,strut=off} 
\floatstyle{ruled}
\newfloat{listing}{tb}{lst}{}
\floatname{listing}{Listing}
\title{From Grounding to Planning: Benchmarking Bottlenecks in Web Agents}
\author{
    Segev Shlomov, Ben wiesel, Aviad Sela, Ido Levy, Liane Galanti, Roy Abitbol
}
\affiliations{
    IBM Research - Israel\\

     University of Haifa Campus, Mount Carmel, Haifa 3498825, Israel \\ 
    \{segev.shlomov1, benwiesel, Ido.Levy1, Liane.Galanti\}@ibm.com, \{sela, roy.abitbol\}@il.ibm.com
}

\usepackage{bibentry}

\begin{document}

\maketitle

\begin{abstract}
General web-based agents are increasingly essential for interacting with complex web environments, yet their performance in real-world web applications remains poor, yielding extremely low accuracy even with state-of-the-art frontier models. We observe that these agents can be decomposed into two primary components: Planning and Grounding. Yet, most existing research treats these agents as black boxes, focusing on end-to-end evaluations which hinder meaningful improvements. We sharpen the distinction between the planning and grounding components and conduct a novel analysis by refining experiments on the Mind2Web dataset. Our work proposes a new benchmark for each of the components separately, identifying the bottlenecks and pain points that limit agent performance. Contrary to prevalent assumptions, our findings suggest that grounding is not a significant bottleneck and can be effectively addressed with current techniques. Instead, the primary challenge lies in the planning component, which is the main source of performance degradation. Through this analysis, we offer new insights and demonstrate practical suggestions for improving the capabilities of web agents, paving the way for more reliable agents.
\end{abstract}

\section{Introduction}
Generalized web agents are designed to autonomously navigate and interact with complex web environments, performing tasks that range from simple information retrieval to intricate multi-step procedures \cite{deng2024mind2web, he2024webvoyager, zhou2023webarena}. As the demand for automation and intelligent interaction with web interfaces grows, these agents are becoming increasingly crucial in various applications, such as virtual assistants, automated customer service, and copilots \cite{drouin2024workarena, zheng2023seeact}.

Significant effort has been dedicated to the development of robust and reliable web agents, ranging from the development of specialized large language models (LLMs) tailored for web navigation \cite{deng2024mind2web, zheng2023seeact} to the incorporation of state-of-the-art (SOTA) generalist models such as GPT-4 with advanced vision capabilities \cite{he2024webvoyager, zheng2023seeact}. Despite these advancements, the performance of web agents in real-world scenarios remains low. These agents frequently fail to achieve reasonable accuracy in completing web-based tasks, raising concerns about their reliability and effectiveness in practical applications \cite{li2024websuite, yoran2024assistantbench}.

A key challenge lies in the prevalent approach to evaluating web agents, which often treats them as black-box systems and focuses on end-to-end performance metrics \cite{deng2024mind2web}. While this approach offers a high-level perspective on agent capabilities, it tends to obscure individual skills and capabilities required for the task, making it difficult to address specific performance issues. In particular, it fails to distinguish between two core components of web agents: Planning and grounding. Planning refers to the agent's ability to determine the appropriate sequence of actions to accomplish a given task, while grounding involves correctly identifying and interacting with relevant web elements based on these decisions \cite{zheng2023seeact}.

Motivated by this understanding, this paper aims to dissect the influence of planning and grounding on web agents' performance. By isolating these factors, our research determines which aspect predominantly affects the agents' ability to complete web-based tasks. This differentiation clarifies pathways for enhancing agent capabilities and provides clearer insights into the optimal allocation of resources within the web-agent's pipeline. To achieve this, we modified the experimental setup of Mind2Web, implementing controlled variations that isolate the planning and grounding components. Specifically, we introduce two distinct modes of operation. The first, termed high-level, aligns with the traditional method of evaluating agents' performance on Mind2Web \cite{deng2024mind2web}. In this mode, the agent is given a general description of a multi-step task and must determine the correct sequence of actions to complete it. The second, termed low-level, gives the model explicit references to the elements it needs to interact with at each step of the task. This mode aims to isolate the grounding component by eliminating planning.


Our experiments reveal several key conclusions. Contrary to previous assertions \cite{zheng2023seeact}, planning emerges as the primary bottleneck limiting overall agent performance. We show that even under controlled setup, where models have to choose between merely two elements, of which one is the ground-truth element, accuracy is much lower than expected. Moreover, we find that when grounding is isolated from planning, a near-perfect element accuracy is achieved, highlighting that focusing only on improving grounding techniques through advanced computer vision (CV) or Document Object Model (DOM)-based methods is not sufficient. Furthermore, we find that reducing the number of candidate elements for the agents consistently improves performance, highlighting the importance of effective element filtration and ranking mechanisms in complex web navigation tasks. Our main contributions are:


\begin{itemize}
    \item We sharpen the decomposition of web-based agents into two core components—Planning and Grounding. This allows for a granular analysis, shifting from black-box evaluation to targeted improvements.
    \item  We introduce a refined benchmark based on Mind2Web, enabling separate evaluation of planning and grounding, revealing that planning is the main bottleneck.
    \item We demonstrate that by improving the page understanding and enhancing the ranking mechanism to support the planning component, our agent, WebNaviX, surpasses the state-of-the-art by 13\%, paving the way for more reliable web agents.
\end{itemize}

\section{Related Work}
\subsubsection{General Web Agents}  Recent advancements have propelled web-based agents to the forefront of navigating and manipulating complex digital environments. These agents are instrumental in a wide array of web automation tasks, including automated testing, personal assistance, data extraction, content management, and enhancing accessibility through assistive technologies \cite{xi2023rise, akter2023depth}. To facilitate the development and benchmarking of these agents, various datasets have been created, such as Mind2Web, AgentBench, and WebArena \cite{liu2023agentbench, zhou2023webarena}. 
These datasets offer a broad spectrum of tasks across multiple domains to test generalizability, evaluate the reasoning and decision-making abilities of agents in interactive environments, and provide a realistic and reproducible web environment.

Current web agents fail to achieve the accuracy required for real-world applications \cite{he2024webvoyager, cheng2024seeclick, wang2024executable, chen2023fireact, pan2024autonomous, koh2024tree}, raising critical questions about their applicability. These agents primarily rely on two components: Planning as the agent's ``mind'', which processes information to formulate subsequent actions, and grounding as the agent's ``eyes'' and ``hands'', which interpret the current state and chosen action to identify and select the correct element. Recent discussions, including claims by \citet{zheng2023seeact}, suggest that models like GPT-4o could effectively plan actions if appropriately grounded. Such insights are crucial as we seek to discern the roles of grounding and planning in enhancing agent efficiency. 

\subsubsection{Grounding} Element grounding is the task of accurately identifying and linking a UI element on a web page, based on a natural language reference name. To do so, the grounder first needs to understand the UI. Semantic UI understanding or as more commonly known, page understanding (PU), plays a critical role in web-based agents, integrating text and image inputs to mimic human capabilities necessary for interacting with graphical user interfaces \cite{carroll2003hci, hegarty2011cognitive, turk2014multimodal}. Originally dominated by HTML-based parsing \cite{deng2022dom, zhou2021simplified}, this field has evolved through the adoption of LLMs that enhance the parsing and comprehension of complex web UIs \cite{gur2022understanding, kim2023language, yin2023lumos, shi2017world}, despite their shortcomings with dynamic and extensive web applications \cite{mitchell2018web}.

Naturally, vision-based soon followed with significant contributions, providing the added benefit of better mimicking the human form of interacting with a web page. Vision-based grounding has evolved from elementary computer vision techniques to more complex models like R-CNN \cite{ren2015faster,manandhar2021magic} and YOLO \cite{redmon2016you,singh2021robust}, which significantly improve the detection and classification of UI elements \cite{chen2020object, xie2020uied}. 

The development of large vision language models (LVLMs) represents a major breakthrough, merging vision and linguistic AI to achieve deeper semantic understanding of UIs \cite{dosovitskiy2020image, ramesh2021zero, radford2021learning}. Recent research confirms the superiority of multi-modal LVLMs in specialized grounding tasks over generalist models \cite{zhang2023reinforced, cheng2024seeclick, lu2024weblinx}.

The integration of text and vision-based grounding with LLMs and LVLMs marks a significant advancement in web agent technology, enhancing the interface between humans and machines in digital environments.
Following the page understanding, the grounder must identify the best match based on the reference element description. This task can be approached using syntactic or semantic matching techniques \cite{ijcai2024p1035}, with some methods leveraging LLMs for improved accuracy \cite{cheng2024seeclick}.

\subsubsection{Planning}
Agent planning has transitioned significantly from its initial stage, where rule-based systems, though useful, were often rigid and limited to predefined tasks \cite{ferrucci2010deepqa}. The advent of LLMs marked a pivotal shift, enabling more dynamic planning across diverse tasks \cite{karpukhin2020dpr, guu2020realm}. A notable advancement was introduced by \citet{nakano2021webgpt}, who utilized LLMs to parse and respond to queries in a text-based web environment, laying the groundwork for more interactive web agents. \citet{gur2024a} further refined this approach by decomposing complex instructions into manageable sub-instructions, enhancing the granularity of task execution. 

The planning evolution has since expanded into two main directions: in-context planning, where agents adjust to tasks within a given context \cite{he2024webvoyager, zhou2023webarena}, and fine-tuned approaches that tailor agent behaviors through specific training regimens \cite{hong2024cogagent}, such as curriculum-based web trajectory learning \cite{lai2024autowebglm}. Despite these innovations, fine-tuned models are expensive to train and to collect high-quality data often struggle with scalability and generalization across the vast and varied web environment. Therefore, their practical implementation in real-life applications is still lacking.

\section{Grounding}
We modified the Mind2Web experimental setup to isolate the action planning from the element grounding. The agent is given a low-level, single-step task and a set of available elements, with the goal of finding (grounding) it by choosing the most suitable element. The Mind2Web dataset was originally planned for evaluating the fulfillment of high-level tasks (e.g., ``Find a 3-bedroom apartment to rent in New York''), using multi-step flows. As such, it does not contain the step-wise instructions serving as the ground truth of each turn in the flow. To that end, we extended Mind2Web so that it is also suitable for evaluating the fulfillment of this low-level task. We used the implicit step-wise data in Mind2Web to extract the action type and the name of the ground truth interacted element for each step in each flow. This results in an augmented dataset containing low-level instructions for each step in Mind2Web.

\subsubsection{Setup} 
The grounding task requires translating low-level instructions---comprising the action type, element name, and a list of available UI elements---into a direct reference to the ground-truth selected element within the MHTML file. To achieve this, we refined the Mind2Web benchmark by creating a subset of 777 cleaned samples, from an initial random sampling of 1000 samples ($~$216 task flows). This subset represents about 10\% of the total test set and preserves the original characteristics of Mind2Web (cross-task: $20.8\%$, cross-web: $15.8\%$, and cross-domain: $63.2\%$) 

Cleaning the benchmark included removing those samples containing duplicate elements that could not be distinguished when using direct references and corrupted samples where the element name was empty or None. Our reasoning is that in those cases, even humans would not be able to distinctively locate the right elements. We note that some of the corrupted samples are due to the inherent process of the Mind2Web annotation, as the ground truth element name was collected automatically by the annotator demonstration on the web page. This refined benchmark enables rapid iteration across numerous experiments, minimizing the costs of commercial frontier LLMs, and reducing experimental time while maintaining the original characteristics of the Mind2Web benchmark.

For a comprehensive evaluation, we employed three types of PU techniques (CV, JS, DOM), two types of prompts (WebVoy and ours), and two core models (GPT-4o and Llama2). We also tested the original MindAct model on this setting. The accuracy metric is calculated according to the element accuracy in \cite{deng2024mind2web}. The entire code base is available at [\url{AnonymouSubmission.git.com}], and the full list of samples is included in the supplementary material.

We create an element selection pipeline with a PU node, syntactic matching node, and semantic matching node. For the PU node, we compare three implementations: a) Vision-based OCR, referred hereafter as CV, b)  Javascript (JS), using  WebVoyager's approach \cite{chen2020object} , and c) our page understanding (PU) algorithm for extracting candidate elements. We refer to it hereafter as DOM PU, indicating that it utilizes the Document Object Model (DOM) analysis techniques. To ensure an accurate comparison, we use the raw HTML bounding box data collected during the original Mind2Web dataset creation, which reflects the elements as they appeared in the original screen view.

\subsubsection{Our}
\label{sec:dom-pu}
DOM-based PU queries a web page using a predefined set of rules, primarily CSS selectors. These rules were hardened following a rigorous process of analysis and measurement, resulting in a high coverage of the interactable elements on the web page. We further enrich each of the elements with semantic information, extracted from the attributes of the queried element, and other DOM elements that are related to it in a spatial or hierarchical connection. This information helps to uniquely characterize each element and will be used by the next nodes in the pipeline. The set of rules can be found in Appendix \ref{sec:appendix_rules}. We note that the set of rules can be easily extended to new applications, and can be even auto-generated by a simple LLM task \cite{ijcai2024p1035}.

The next pipeline step ranks the extracted candidates using common syntactic matching algorithms \cite{levenshtein1966binary}. We match the step-level instruction with the semantic attributes of the candidate elements. Following the syntactic matching, we invoke semantic ranking using a sentence transformer (all-mpnet-base-v2). We sort the elements based on the similarity of their embedding against the step-level instruction. Finally, a (v)LLM uses the entire information to select the appropriate element ID for invocation.
Aligned with findings in \cite{chen2020object}, we also implemented a pipeline primarily based on CV techniques, combined with an optical character recognition (OCR) model \cite{rotman2022detection}. This pipeline ingests a capture of a GUI screen (from the Mind2Web subset) and a text used as the reference expression (e.g., ``Click Submit'' or ``Type First Name''). We match the extracted text and assign each element a score \cite{levenshtein1966binary}. We further utilize a secondary match that considers neighboring elements, based on the assumption that semantic context may stem from nearby elements. Finally, a target click position (x, y coordinate) is generated as output.

\begin{figure}[ht!]
    \centering
    \includegraphics[width=\linewidth]{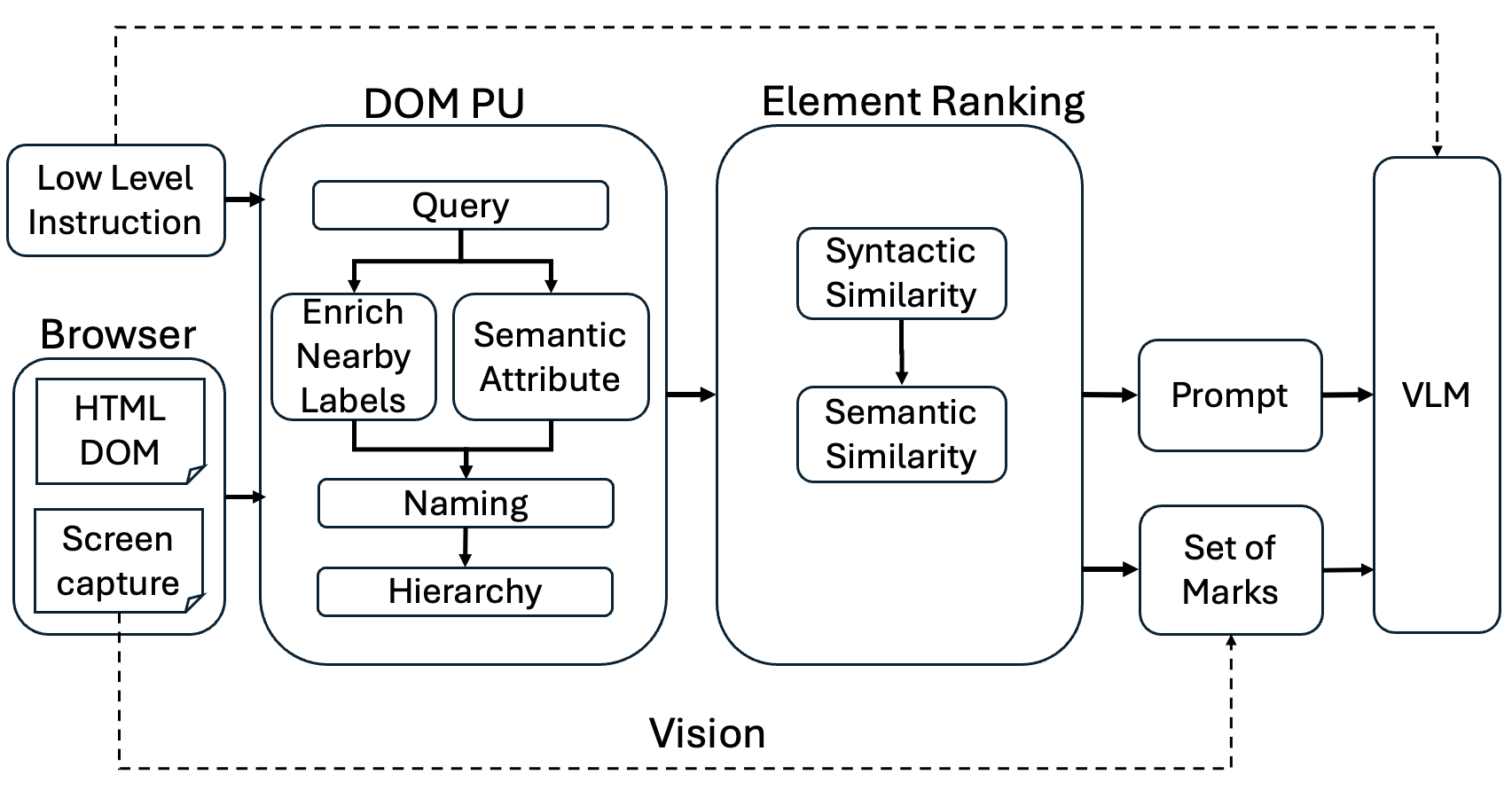}
    \caption{Grounding Pipeline: The website's DOM, screenshot, and low-level instruction are processed through the PU and element ranking phases to generate an annotated screen capture with SoM and a prompt. These are then passed together with the low-level instruction to the VLM to select the web element}
    \label{fig:enter-label}
\end{figure}

\subsubsection{WebVoyager}
\label{sec:WebVoy}
Building on the work in \cite{he2024webvoyager}, we utilized WebVoyager's (WebVoy) code to create an additional experimental pipeline. WebVoy employs a generalist model for both high-level task planning and low-level element grounding. Although originally designed for online applications, we tested WebVoy using the provided code in an offline scenario with the Mind2Web dataset, a context where it had not been previously applied. To address this, we made minimal modifications to the original code to simulate an online interaction, but we preserved the core thought and action prompting as in the original paper. For simplicity and uniformity with other experiments, we reduce the set of permitted actions to click, type, and scroll. Scrolling is emulated by passing different sections of the web page corresponding to the model's scrolling actions. This approach enables us to evaluate performance on Mind2Web while preserving the core methodology of WebVoy.

Initially, WebVoy employs a set-of-marks (SoM) algorithm \cite{yang2023set} to identify candidate UI elements on the screen, drawing boxes around them, and assigning an ID to each element. The processed image and the textual list of elements are then used in the subsequent phase. For the SoM, WebVoy uses a simple JavaScript (JS) query to extract all interactive elements from the web page. After the SoM step, WebVoy creates a prompt for a LVLM (capable of ingesting both text and image, GPT-4o in our setup) which includes the processed SoM image, the list of SoM candidates, and a low-level instruction. Leveraging its visual reasoning capabilities, GPT-4o processes this input data and identifies the ID of the appropriate element. WebVoy completes the step by either interacting with the element or, in offline evaluations, comparing the result to the ground truth.

\subsection{Results}
As illustrated in Table \ref{tbl:grounding}, both our text-only DOM-based algorithm and the GPT-4o vision-based method surpassed 85\% accuracy in element grounding. Specifically, our DOM-based algorithm achieved a 90.4\% success rate (703 out of 777) in accurately identifying the correct element, indicating that element grounding is not a significant bottleneck. Additionally, the inclusion of computer vision techniques or a large vision language model resulted in only a slight improvement in accuracy. For instance, when vision was integrated, the accuracy increased marginally from 77.6\% to 78.4\% on the JS PU, and from 88.3\% to 90.4\% with our DOM-based PU approach. We attribute this to the inherently well-defined nature of the grounding task, where the primary requirement is to identify the correct reference element. In this context, a visual representation of the entire user interface does not offer substantial additional benefits. Furthermore, there is minimal variation in the base model performance when using a text-only LLM. For example, Lamma2 achieved 87.1\% accuracy, while GPT-4o reached 88.3\%. We also evaluated a computer vision-only approach, which yielded a relatively lower accuracy of 70.2\%. This limitation is consistent with challenges in OCR models \cite{xie2020uied,qian2022accelerating}, as they often fail to detect all UI elements on the screen and lack access to crucial metadata (e.g., element type, Aria label, and element role).

\begin{table}[!ht]
    \centering
    \resizebox{0.98\linewidth}{!}{
    \begin{tabular}{lcccc}
        \addlinespace[2pt] 
        \toprule
        \textbf{Method} & \textbf{Model} & \textbf{Vision} & \textbf{PU} & \textbf{Accuracy} \\[0.5mm]
        \midrule
        MindAct & GPT-4o & N & DeBERTa & 87 \\ 
        WebVoy & GPT-4o & N & JS & 77.6\\ 
        Our & Llama2 70B & N & Our Dom-PU & 87.1 \\
        Our & GPT-4o & N & Our Dom-PU & 88.3 \\ 
        \midrule
        WebVoy & GPT-4o & Y & JS & 78.4 \\
        Our & GPT-4o & Y & CV & 70.2 \\
        Our & GPT-4o & Y & Our Dom-PU & \textbf{90.4} \\
    \end{tabular}}
    \caption{Element accuracy results of different methods over the benchmark dataset on the low-level grounding task.}
    \label{tbl:grounding}
\end{table}

We further manually analyzed the remaining 9.6\% of errors and identified that approximately half of them stem from inherent issues within the Mind2Web dataset itself. One common issue involves cases where the task instructs the agent to ``click'' on a typeable object, such as a text input field, where a typing action would be more appropriate. Additionally, some instructions within the dataset are too vague, lacking sufficient detail to precisely identify the target element. For example, instructions like ``Click on 5'' do not always specify which attribute or visible text to match, leaving ambiguity that the grounding algorithms struggle to resolve. Another frequent source of error arises from elements nested within other elements, where the ground truth annotation only highlights the inner or outer element, leading to difficulties in accurately identifying the entire interactive area. These issues are compounded by the fact that the annotations in Mind2Web were created through demonstrations, rather than reflecting how an agent or user would naturally reference the element in a real-world scenario, which can result in subtle discrepancies that impact grounding accuracy. The remaining half of the errors can be attributed to limitations in our current PU algorithmic approach. Examples can be found in Appendix \ref{sec:grd-err-anls}.

\section{Planning}

Going back to the original settings of Mind2Web, the high-level objective setting is the standard setting of web agents, where a natural language description of the intent is given (e.g., ``Order a United flight from NY to SF on Aug 8'') and the agent should act one step at a time and perform the actions required to accomplish the objective. Typically, each step's input includes the current web page, the past action commands, and the main objective. The output is the type of UI action and a reference (element id) to the element that the agents would like to interact with to accomplish its goal. For analyzing action planning, we use only the subset of 703 successfully grounded samples as a baseline to evaluate planning decisions independently, minimizing grounding-related challenges effect. We tested four models (MindAct, SeeAct, WebVoy, and ours - WebNaviX) with three PU types (JS, DeBERTa, and ours). To ensure a fair comparison and validate the efficiency of the split, we employed two mechanisms: comparing the weighted average results of SOTA models on the original Mind2Web split and projecting their performance onto our split (cross-task: 31.1\%, cross-web: 11.52\%, and cross-domain: 57.3\%).

\subsubsection{MindAct}
To ensure consistency, we ground the results on our benchmark dataset split by applying SOTA web navigation methods as a baseline. The first method we use is MindAct \cite{deng2024mind2web}, which implements a two-stage, text-only web navigation model consisting of candidate generation and action prediction stages. In the first stage, MindAct employs a ranker to select the top 50 elements. Subsequently, the action generation problem is framed as a multiple-choice question-answering task, with the candidate elements serving as options, including a ``None'' option if the target element is absent. We run the provided code, initially producing ranking results using their DeBERTa model (trained on the entire Mind2Web training dataset). And later, we run either their fine-tuned Flan-T5XL model weights or an in-context learning GPT-4o model for action prediction. 

\subsubsection{SeeAct}
We include results from SeeAct \cite{zheng2023seeact} to provide context for our large vision-language approach. However, since the provided SeeAct code is only suitable for online evaluation, we project the performance on the offline dataset by calculating a weighted average of the respective results on cross-task, cross-website, and cross-domain using their relative presence on the new split. 

\subsubsection{WebVoyager}
 We used the same method as described in Section \ref{sec:WebVoy}. However, as opposed to the grounding experiment, in this planning setup, we prompt the LLM with the high-level text describing the task. 

\subsubsection{WebNaviX} 
Our agent's first step is applying the DOM-Based PU as described in the grounding task (Section \ref{sec:dom-pu}). Thereafter, the results of the DOM-PU are ranked using semantic similarity (as discussed in Section \ref{sec:dom-pu}). The resulting elements are then transformed into a SoM list of candidates and embedded (like in WebVoyager) in the prompt of the LVLM together with the  history of past actions, and the high-level task description. The LVLM planner returns the predicted action: Type or click, and the predicted element ID. If the action is type, the planner also provides the required value for typing in. 

\begin{figure}[ht!]
    \centering
    \includegraphics[width=0.99\linewidth]{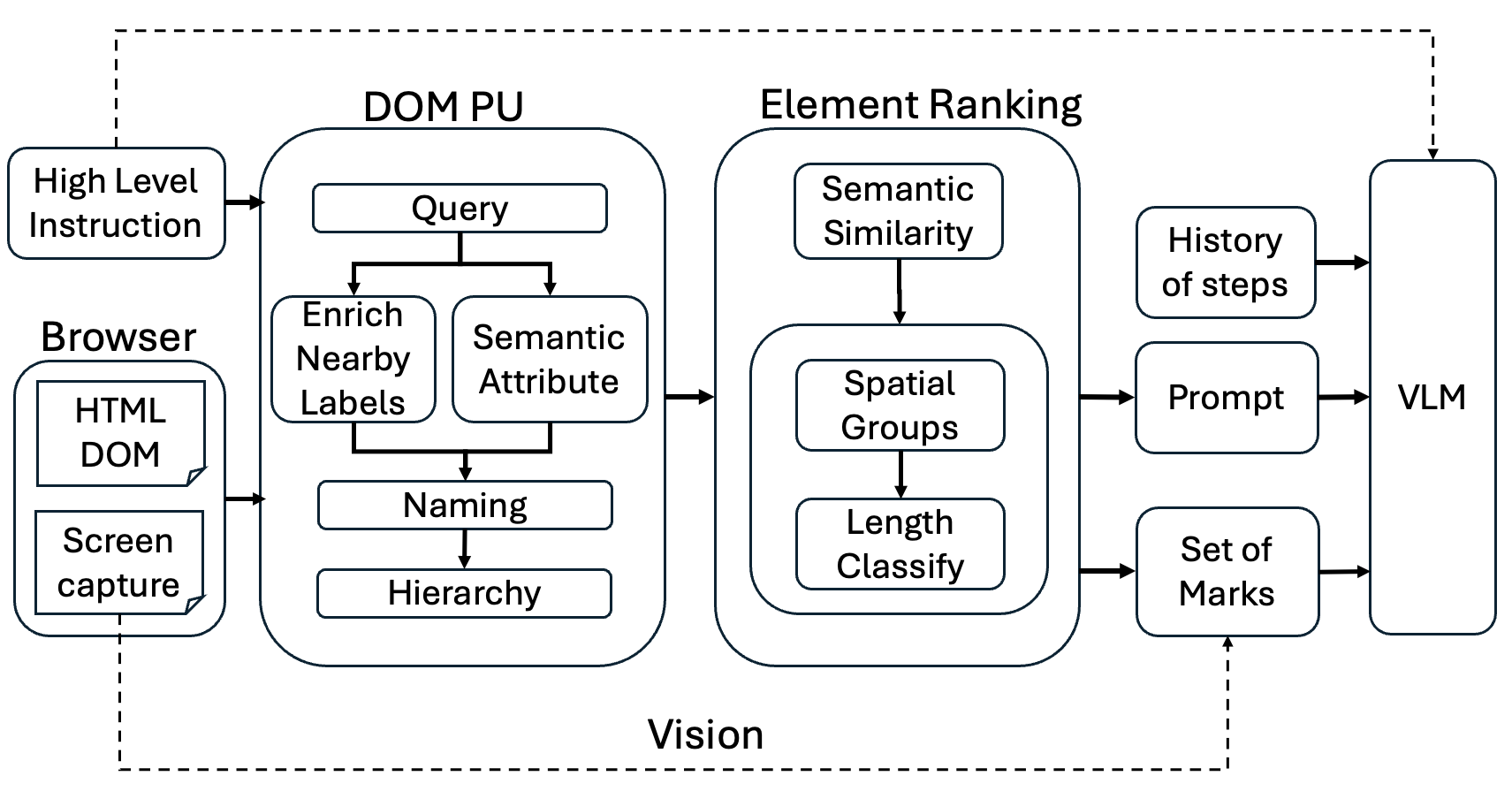}
    \caption{WebNaviX Architecture: The website's DOM, screenshot, and high-level instruction are processed through the PU and element ranking phase equipped with an improved ranking to generate an annotated screen capture with SoM and a prompt. These are then passed together with the high-level instruction and histroy of actions to the VLM to select the web element.}
    \label{fig:architecture-high}
\end{figure}

\subsection{Results}

Table \ref{tbl:base} presents the results of action planning using various approaches. We focus solely on element accuracy, omitting the Operation-F1 and Step Success Rate (Step-SR) metrics used in \cite{deng2024mind2web}. Our analysis reveals that selecting the correct operation is relatively straightforward and can often be inferred from the type of the selected element. Therefore, we consider the element accuracy as the most significant metric that directly correlates with the Step-SR.

\begin{table}[!ht]
    \centering
    \resizebox{\linewidth}{!}{
    \begin{tabular}{lcccc}
        \addlinespace[2pt] 
        \toprule
        \textbf{Base Method} & \textbf{Model} & \textbf{Vision} & \textbf{PU} & \textbf{Accuracy} \\[0.5mm] 
        \midrule
        \multicolumn{3}{l}{\textbf{Supervised Fine-Tuning}} \\
        \cmidrule(lr){1-2}
        \addlinespace[2pt]
        MindAct & Flan-T5L & N & DeBERTa & 44.10 \\ 
        MindAct & Flan-T5XL & N & DeBERTa & 46.75 \\ 
        \midrule
        \multicolumn{3}{l}{\textbf{In-Context Learning}} \\
        \cmidrule(lr){1-2}
        MindAct & GPT-4o & N & DeBERTa & 36.27 \\ 
        WebVoy & GPT-4o & N & JS & 27 \\ 
        WebNaviX (our) & Lamma3 70B & N & Our-PU & 35.13\\
        WebNaviX (our) & GPT-4o & N & Our-PU & \textbf{40.68}$^*$\\
        \midrule
        WebVoy & GPT-4o & Y & JS & 32.86 \\ 
        SeeAct & GPT-4o & Y & DeBERTa & 43.1 \\ 
        WebNaviX (our) & GPT-4o & Y & Our-PU & 47.37 \\
        WebNaviX +R (our) & GPT-4o & Y & Our-PU & \textbf{49.08$^*$} \\
    \end{tabular}}
    \caption{Element accuracy results of different methods over the benchmark dataset in high-level setting. WebNaviX significantly outperforms SOTA methods. Vision techniques are essential for in-context web agents. The supervised methods involve fine-tuning of the underlying model.}
    \label{tbl:base}
\end{table}

To ensure the validity of our experiments, we compare the performance of SOTA models on our benchmark split with their original results. On our split, MindAct FlanT5XL projects an accuracy of $46.53\%$, calculated as a weighted average across cross-task ($31.1\%$), cross-web ($11.52\%$), and cross-domain ($57.3\%$) subsets. The actual result on our split was $46.75\%$. MindAct GPT4 projects an accuracy of $36.45\%$, where the actual result on our split is $36.27\%$, showing insignificant differences from the projection.

As shown in Table \ref{tbl:base}, the current SOTA MindAct FlanT5 model demonstrates strong performance on our Mind2Web split. This is primarily due to our split containing more cross-task content, which plays to the strengths of fine-tuned models. However, fine-tuned models show reduced effectiveness in cross-website and cross-domain evaluations. Furthermore, the process of fine-tuning models for specific tasks, websites, or domains requires large training data and can be resource-intensive, potentially limiting their practical applicability and generalizability.

In contrast, our analysis of in-context learning models reveals that for text-based methods (Vision=N), our agent WebNaviX, utilizing the DOM-based PU technique with the ranking mechanism and the WebVoyager's prompt, yields the best result (40.68\%). Notably, it outperforms the JS technique from WebVoy, demonstrating the efficacy of our PU method. For vision-based models, our agent WebNaviX achieves significant (Wilcoxon Signed-Rank test \cite{dror2018hitchhiker, dror2020statistical}, $p<0.05$) improvement over existing state-of-the-art methods without relying on data-specific fine-tuning.
Using our method allows you to avoid task-specific fine-tuning while maintaining consistent performance across diverse splits, thereby addressing the key limitation of previous approaches.

\subsubsection{Planning is the main bottleneck}

A critical challenge faced by LLMs in web-based agents is their ability to effectively plan actions based on a list of relevant UI elements. The planner component, which is responsible for selecting the next element to interact with, frequently struggles to make accurate decisions, even when provided with a narrowed list of options. To further investigate this issue, we conduct a series of experiments where the ground truth element is explicitly injected into the list of choices presented to the planner (Oracle Ranker). 
We tested the planner's performance by varying the number of relevant UI elements, focusing on its ability to correctly identify the ground truth when it is guaranteed to be included in its list of options. This manipulation helps isolate the impact of the planning algorithm from grounding or element ranking challenges.

The result, as can be seen in the Oracle Ranker column on Table \ref{tab:ranking_comparison}, reveals a surprising and concerning trend. Even when the planner is tasked with choosing between just two elements, one of which is the ground truth, it only achieves an accuracy of 86\%. Given the narrow decision space, we consider this to be a low rate of accuracy. It serves as a testament to the fact that the planner struggles to make the right choice, even under a simplified setup, let alone under more complex conditions. These findings suggest that the current planning mechanisms within LLMs are insufficient for achieving the high accuracy required in complex web environments. Contrary to the paper \cite{zheng2023seeact} titled ``GPT-4V(ision) is a Generalist Web Agent, if Grounded'', our results show that currently LVLMs cannot act as web agents, even if perfectly grounded. We argue that additional external knowledge may be necessary to enhance the planner's capabilities. Without such improvements, the planning phase remains a substantial bottleneck, limiting the overall effectiveness of web-based agents. An example that shows the difficulty of the planner can be found in Appendix \ref{sec:grd-err-anls}.

\begin{figure}[ht!]
\centering
\includegraphics[width=0.9\linewidth]{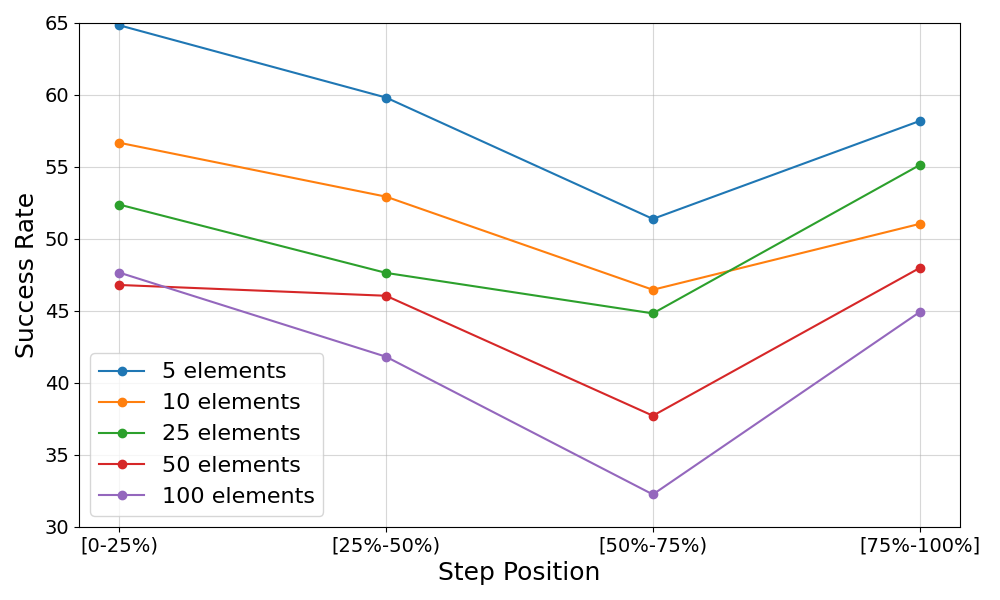}
\caption{Planner performance across task flow steps.}
\label{fig:step_pos}
\end{figure}

We further conducted an analysis of the planner's performance by examining how its accuracy varied across different step positions within the task flow. We initially hypothesized that the accuracy might be low in the early steps, since there are multiple correct ways to initiate tasks but only one is annotated as ground truth. In the final steps, where the planner would have accumulated enough context from previous steps, we expected high performance. As shown in Figure \ref{fig:step_pos},
the accuracy forms a U-shaped pattern, with better performance at the start and end of the task flow, and a small dip in the middle. This consistent behavior across multiple ranking numbers suggests that mid-task complexity and variability pose more challenges for the planner.

\subsection{Improving Ranking Heuristics}

To enhance the candidate generation process, we explored additional ranking heuristics beyond semantic similarity. We observed that most ground truth candidates in the Mind2Web training dataset are short, with 92\% containing less than 6 words. To counter this bias, we divided the candidates into three groups: up to 3 words, 4-6 words, and 7+ words. Elements were then sampled from each group in descending order of semantic similarity, while enforcing a sampling distribution corresponding to the natural distribution of text lengths. This approach improved the recall of ground truth elements (see the +Length row in Table \ref{tab:ranking_methods}).

\subsubsection{Location-based Heuristic}
\begin{figure}[ht!]
    \centering
    \includegraphics[width=0.9\linewidth]{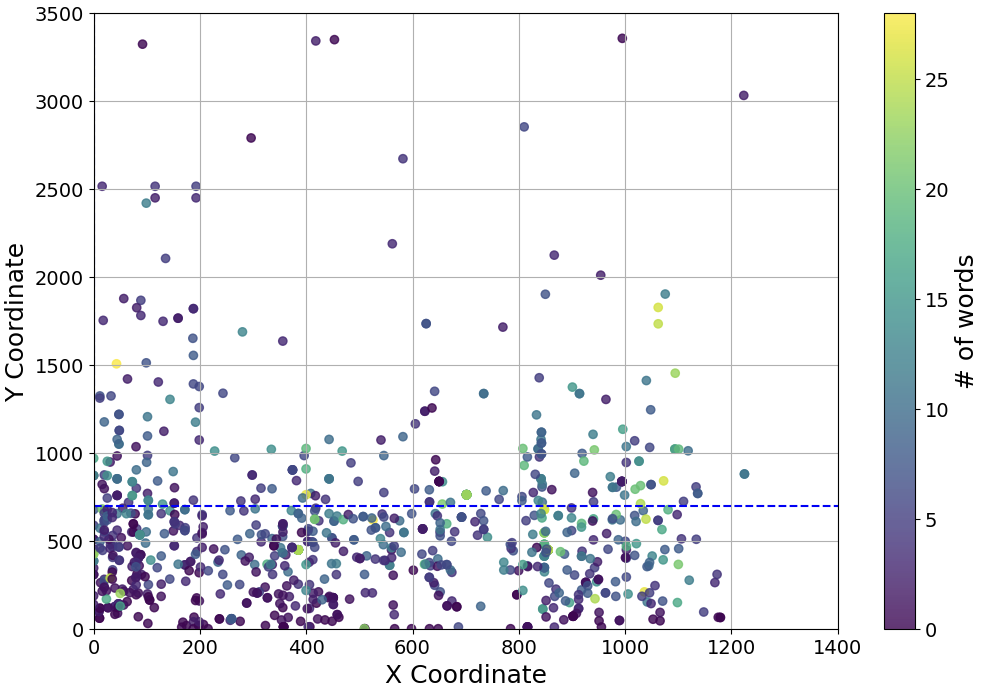}
    \caption{Spatial distribution of the ground truth element relative to the page layouts, within the training dataset. Lower Y values indicate proximity to the top of the page and lower X values indicate proximity to the left side of the page. Colors indicate the length of the text.}
    \label{fig:gt-spatial}
\end{figure}

We observed that the ground truth elements tend to cluster towards the top and left sides of pages (Figure \ref{fig:gt-spatial}). To address this spatial bias, we split candidates into two groups based on their Y-axis positions: Up to 700 pixels and above 700 pixels. We then implemented an over-sampling strategy for elements from the 0-700 px group with a 0.92 sampling ratio. Combining these length and location strategies yielded further improvements in recall rates (see the +Length+Location row in Table \ref{tab:ranking_methods}).

\subsubsection{Performance Comparison}

Table \ref{tab:ranking_comparison} illustrates the substantial impact of the improved ranker on overall performance. The +Length+Location method consistently outperforms the semantic only approach across all candidate set sizes, with the optimal set size emerging as 30 candidates. Notably, the +Length+Location method exhibits a more plateaued pattern, suggesting early identification of relevant candidates and reducing the need for large candidate sets.
While the fine-tuned DeBERTa-based ranker from MindAct slightly outperformed our approach, our ranker offers significant advantages in flexibility and generalizability, requiring no specific fine-tuning. This characteristic makes our approach adaptable to diverse datasets and domains, potentially offering superior performance in rapidly evolving real-world applications.

\begin{table}[ht!]
  \centering
  \small
  \begin{tabular}{lcccc}
    \toprule
    \textbf{Ranking Method} & \multicolumn{4}{c}{\textbf{Number of Candidates}} \\
    \cmidrule(lr){2-5}
    & \textbf{20} & \textbf{30} & \textbf{50} & \textbf{70} \\
    \midrule
    Semantic only & 54 & 64 & 74 & 80 \\
    +Length & 62 & 69 & 80 & 85 \\
    +Location & 72 & 80 & 93 & 95 \\
    +Length+Location & 73 & 83 & 92 & 96 \\
    +Location+Step & 72 & 81 & 94 & 97 \\
    +Length+Location+Step & 75 & 85 & 94 & 97 \\ 
    \addlinespace[0.3em]
    \hdashline[2pt/2pt]
    \addlinespace[0.3em]
    MindAct & 84.6 & -- & 93.8 & -- \\
    \bottomrule
  \end{tabular}
  \caption{Recall rates \% of different ranking methods across varying numbers of candidates.}
  \label{tab:ranking_methods}
\end{table}
\begin{table}[htbp]
\centering
\small
\setlength{\tabcolsep}{4pt}
\begin{tabular}{cccc}
\toprule
\textbf{Candidates} & \textbf{Semantic} & \textbf{+Length} & \textbf{Oracle} \\
\textbf{Number} & \textbf{Only} & \textbf{+Location} & \textbf{Ranker} \\
\midrule
2 & -- & -- & 86.05 \\
5 & -- & -- & 75.25 \\
10 & 32.43 & 43.67 & 70.41 \\
20 & 38.55 & 47.94 & 63.72 \\
30 & 40.11 & \textbf{49.08} & 62.30 \\
50 & 43.10 & 48.65 & 57.75 \\
All & 47.37 & 47.37 & 47.37 \\
\bottomrule
\end{tabular}
\caption{Element accuracy comparison using different ranking methods, across different numbers of candidates. The Oracle ranker column corresponds to WebNaviX result with the ground true injected into the list of UI elements.}
\label{tab:ranking_comparison}
\end{table}
\section{Discussion}

We isolate planning and grounding components by conducting controlled experiments on a new refined Mind2Web benchmark. Our results indicate that the primary bottleneck is the model's planning capabilities. We believe that simply using a "smarter" model, more computational power, or a more advanced frontier model, or even improving grounding to perfection, will have a limited impact on accuracy in real-world web agents. Instead, we hypothesize that the main gap lies in external knowledge. Incorporating additional knowledge and context into the planner, particularly in business applications, could bridge the gap between abstract task descriptions and concrete, step-by-step actions.

Building on this premise, we experimented with a simple modification to our algorithm: a single call to an LLM at the outset of each workflow. This LLM is tasked with generating a high-level pre-plan for task completion, constrained to a concise paragraph (see Appendix \ref{appendix:preplanning} for more details). This pre-plan is then appended to the prompt of each step in the flow when the LLM is called upon to select the optimal candidate. Using this simple method, the performance improved marginally (51.7\%). This improvement, although not significant,  suggests the potential benefits of embedding additional context in planning decisions and hints at the value of a broader strategic overview guiding action sequences. Further research is needed to fully understand the implications and reliability of this approach.

\subsubsection {Threats to Validity} 
We selected Mind2Web for our experiments because it is the first widely adopted large-scale dataset derived from real websites, providing a diverse and representative set of tasks for real-world web interactions. However, it is important to acknowledge our limitations.

We did not perform an exhaustive hyperparameter optimization. However, this aligns with our research objectives, which prioritize assessing fundamental planning and reasoning capabilities over achieving peak performance. Although we were able to beat the SOTA models this is not the main goal of the paper. We selected and rigorously tested SOTA models for their reasoning abilities, providing a strong baseline and realistic results. We argue that if these frontier models show low planning performance, it is likely that other models would struggle even more.
In addition, while fine-tuned SOTA models may achieve performance close to our technique, they require training data and struggle with cross-domain tasks. Although these methods can boost performance in specific contexts, their limitations highlight the need for more generalized approaches that can adapt across diverse domains.

Our study is based on offline experiments, which may not fully capture the dynamic nature of live web environments. This discrepancy could lead to potential differences between our experimental results and real-world performance. To mitigate this limitation, we carefully adapted algorithms to function effectively in an offline setting, striving to maintain the integrity of the original methods. However, we acknowledge that these adaptations may not entirely replicate the complexities of live environments. Additionally, we chose not to test against environments like WebArena, as they conflate planning and grounding tasks. While such datasets are valuable for overall agent assessment, our focus on isolating grounding from planning was critical for pinpointing the root causes of performance deficits in web interaction tasks.

Our study also uses only a subset of the Mind2Web dataset potentially limiting the generalizability of our findings. To mitigate this, we carefully selected a random subset, and despite removing duplicates and corrupt samples, we maintained the original distribution characteristics of Mind2Web. This approach ensures that our results remain representative of the full dataset, as evidenced by the consistency of results across multiple algorithms and models. Moreover, we projected the results of the baseline algorithms on our subset to accumulate fair comparison. We also focused on common web interactions (e.g., Click and Type commands) and did not explore tasks like copy-pasting or data extraction, leaving these for future research.




\bibliography{aaai25}

\section*{Reproducibility Checklist}
\textbf{This paper:}
\begin{itemize}[label=$\circ$]
    \item Includes a conceptual outline and/or pseudocode description of AI methods introduced (\textbf{yes})
    \item Clearly delineates statements that are opinions, hypothesis, and speculation from objective facts and results (\textbf{yes})
    \item Provides well marked pedagogical references for less-familiare readers to gain background necessary to replicate the paper (\textbf{yes})
\end{itemize}

\noindent\textbf{Does this paper make theoretical contributions? (no)}\\
If yes, please complete the list below.
\begin{itemize}[label=$\circ$]
    \item All assumptions and restrictions are stated clearly and formally. (yes/partial/no)
    \item All novel claims are stated formally (e.g., in theorem statements). (yes/partial/no)
    \item Proofs of all novel claims are included. (yes/partial/no)
    \item Proof sketches or intuitions are given for complex and/or novel results. (yes/partial/no)
    \item Appropriate citations to theoretical tools used are given. (yes/partial/no)
    \item All theoretical claims are demonstrated empirically to hold. (yes/partial/no/NA)
    \item All experimental code used to eliminate or disprove claims is included. (yes/no/NA)
\end{itemize}

\noindent\noindent\textbf{Does this paper rely on one or more datasets? (yes)}\\
If yes, please complete the list below.
\begin{itemize}[label=$\circ$]
    \item A motivation is given for why the experiments are conducted on the selected datasets (yes)
    \item All novel datasets introduced in this paper are included in a data appendix. (yes)
    \item All novel datasets introduced in this paper will be made publicly available upon publication of the paper with a license that allows free usage for research purposes. (yes)
    \item All datasets drawn from the existing literature (potentially including authors’ own previously published work) are accompanied by appropriate citations. (yes)
    \item All datasets drawn from the existing literature (potentially including authors’ own previously published work) are publicly available. (yes)
    \item All datasets that are not publicly available are described in detail, with explanation why publicly available alternatives are not scientifically satisficing. (NA)
\end{itemize}

\noindent\textbf{Does this paper include computational experiments? (yes)}
If yes, please complete the list below.
\begin{itemize}[label=$\circ$]
    \item Any code required for pre-processing data is included in the appendix. (yes).
    \item All source code required for conducting and analyzing the experiments is included in a code appendix. (yes)
    \item All source code required for conducting and analyzing the experiments will be made publicly available upon publication of the paper with a license that allows free usage for research purposes. (yes)
    \item All source code implementing new methods have comments detailing the implementation, with references to the paper where each step comes from (yes)
    \item If an algorithm depends on randomness, then the method used for setting seeds is described in a way sufficient to allow replication of results. (yes)
    \item This paper specifies the computing infrastructure used for running experiments (hardware and software), including GPU/CPU models; amount of memory; operating system; names and versions of relevant software libraries and frameworks. (yes)
    \item This paper formally describes evaluation metrics used and explains the motivation for choosing these metrics. (yes)
    \item This paper states the number of algorithm runs used to compute each reported result. (yes)
    \item Analysis of experiments goes beyond single-dimensional summaries of performance (e.g., average; median) to include measures of variation, confidence, or other distributional information. (yes)
    \item The significance of any improvement or decrease in performance is judged using appropriate statistical tests (e.g., Wilcoxon signed-rank). (yes)
    \item This paper lists all final (hyper-)parameters used for each model/algorithm in the paper’s experiments. (yes)
    \item This paper states the number and range of values tried per (hyper-) parameter during development of the paper, along with the criterion used for selecting the final parameter setting. (yes)
\end{itemize}

\appendix

\section{Mitigating Text Length Bias in Semantic Matching}
Our analysis revealed a bias in the LLM's candidate selection towards elements with longer texts. We hypothesize that this bias stems from the typically lengthy high-level task descriptions (usually at least one sentence), leading the naive semantic similarity ranker to favor candidates with longer text. However, the majority of ground truth candidates on the original Mind2Web training dataset are actually quite short, with 92\% of the samples containing less than 6 words.

To address this discrepancy, we introduced a counter-bias mechanism. We divided the candidates into three groups based on their text length: up to 3 words, 4-6 words, and 7+ words. When generating candidates, we sample elements from each group in descending order of semantic similarity, while enforcing a sampling distribution that corresponds to the natural distribution of text lengths. This approach effectively up-samples shorter words.

The efficacy of this method is evident in the results. The "+Length" ranking method shows a significant improvement in the recall of ground truth elements (i.e., the percentage of times the ground truth element was included in the candidates list). For instance, with 50 candidates, the accuracy improves from 74\% (Semantic only) to 80\% (+Length). This improvement is consistent across different numbers of candidates, demonstrating the robustness of our length-aware sampling approach.

\begin{figure}[ht!]
    \centering
    \includegraphics[width=0.9\linewidth]{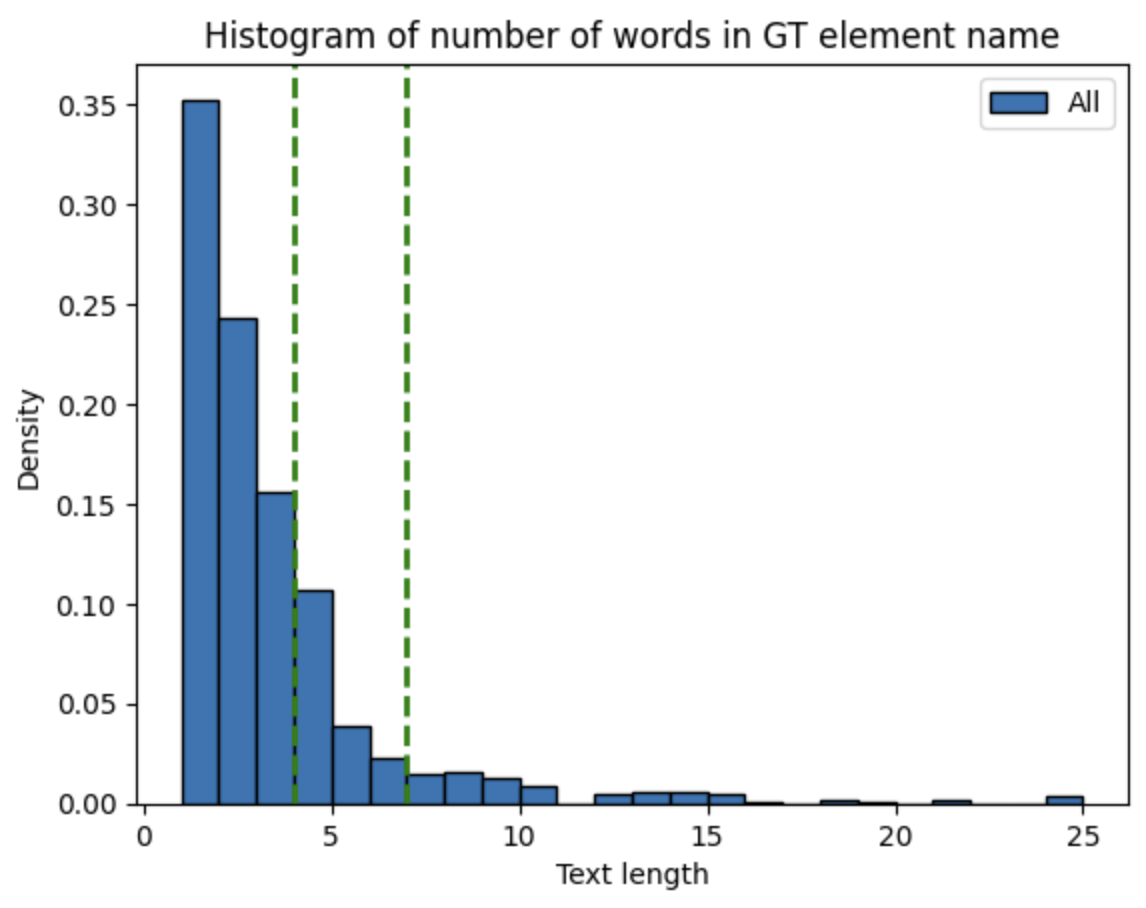}
    \caption{Frequency histogram of the number of words in the ground truth text on the Mind2Web training dataset.}
    \label{fig:gt-len}
\end{figure}

\section{Element Grounding Using DOM Parsing}
\label{sec:appendix_rules}

\begin{figure}[ht!]
  \centering
  \includegraphics[width=\linewidth]{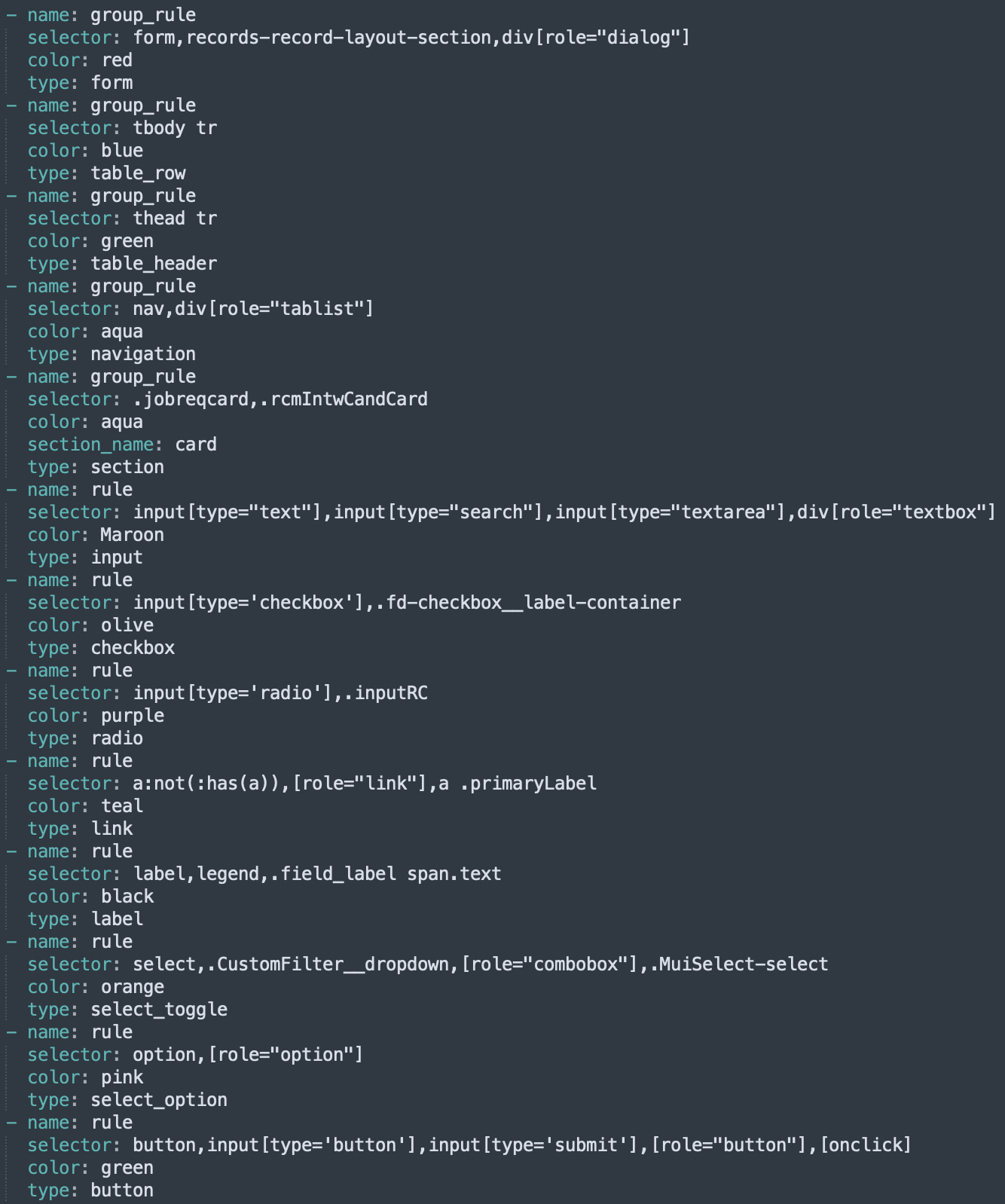}
  \caption{CSS Selector rules used for DOM-based web element grounding}
  \label{fig:rules}
\end{figure}

To facilitate robust web element grounding, we utilized a set of CSS Selector rules, outlined in Figure \ref{fig:rules}. These rules target specific DOM elements, enhancing the accuracy of element identification during DOM-based parsing operations. Each rule defines a unique CSS Selector, ensuring high coverage across varied web pages and applications. Below we describe key rules employed in our experiments:

\begin{itemize}
  \item \textbf{Form Elements:} Target forms with complex layouts or modal dialogs, colored in red for critical interaction points (\texttt{form, .records--layout-section, div[role=``dialog'']}).
  \item \textbf{Table Rows:} Blue-coded selectors (\texttt{tbody tr}) to parse data tables efficiently.
  \item \textbf{Navigation Tabs:} Aqua-colored selectors for navigation elements (\texttt{nav[role=``tablist'']}).
  \item \textbf{Input Fields:} Various inputs including text, search, and textarea are highlighted with maroon, crucial for form interactions (\texttt{input[type=``text''], input[type=``search''], input[type=``textarea'']}).
  \item \textbf{Selection Controls:} Checkbox and radio inputs are distinctly styled in olive and purple, respectively, for clear visibility and interaction (\texttt{input[type='checkbox'],} \texttt{input[type='radio']}).
  \item \textbf{Links and Buttons:} Non-advertorial links and actionable buttons are tagged in teal and green, promoting straightforward navigation and actions (\texttt{a:not(:has(img)),}\texttt{button}).
\end{itemize}

These rules ensure comprehensive coverage and precise targeting within a DOM, facilitating the subsequent syntactic and semantic matching processes crucial for effective element grounding.

\section{Hyperparameters} \label{appendix:hyperparameters}

\subsection{Setting of Hyperparameters}
We employed a variety of models with specific hyperparameters optimized for each task. For the \texttt{GPT-4o} model on the OpenAI API, parameters were set as follows: \texttt{max\_tokens} to 128K, \texttt{temperature} and \texttt{top\_p} both at 1, with \texttt{frequency\_penalty} and \texttt{presence\_penalty} at 0. We used a \texttt{batch\_size} of \texttt{1}, and configured the API calls to allow up to \texttt{10} retries with a timeout of \texttt{60} seconds.

\subsection{DOM Grounding and Planning}
For the DOM Grounding (Dom-PU) task, we used the \texttt{llama-2-70b} model, specifying a \texttt{greedy} decoding method, \texttt{temperature} of 0.1, and a maximum of 20 new tokens. DOM Planning utilized \texttt{meta-llama/llama-3-70b-instruct} with a token range of 1 to 200. Both settings employed policies with a threshold of 0.75, reflecting input and output considerations.

Hyperparameter selection was based on the criterion of maximizing performance results. Despite the high cost of experimentation, which restricted extensive parameter tuning, there is a wide range and number of values tested.

\section{MIND2WEB Dataset}
\label{sec:dataset}
MIND2WEB (\url{https://osu-nlp-group.github.io/Mind2Web/} dataset \cite{deng2024mind2web}) serves as a benchmark for the development of generalist agents that interpret and execute language-driven instructions across a spectrum of real-world websites. It composed of 2,350 open-ended tasks sourced from 137 websites across 31 diverse domains such as travel, shopping, and services. It stands out by employing actual web environments rather than simulations, capturing the complexity of modern interfaces and user interactions. 

Tasks range from simple queries to complex sequences that require navigation, transaction, or data analysis, providing a rich tapestry of user interactions. Each task is accompanied by natural language descriptions, annotated sequences of actions, and comprehensive web page snapshots—HTML, DOM trees, screenshots, and network traffic—documenting each step.

The generalization capability of agents is tested through an evaluation framework that includes cross-task, cross-website, and cross-domain challenges. This framework assesses agents' ability to adapt to unseen tasks, new websites within familiar domains, and completely novel domains.

\begin{figure*}[ht!]
  \centering
  \includegraphics[width=\linewidth]{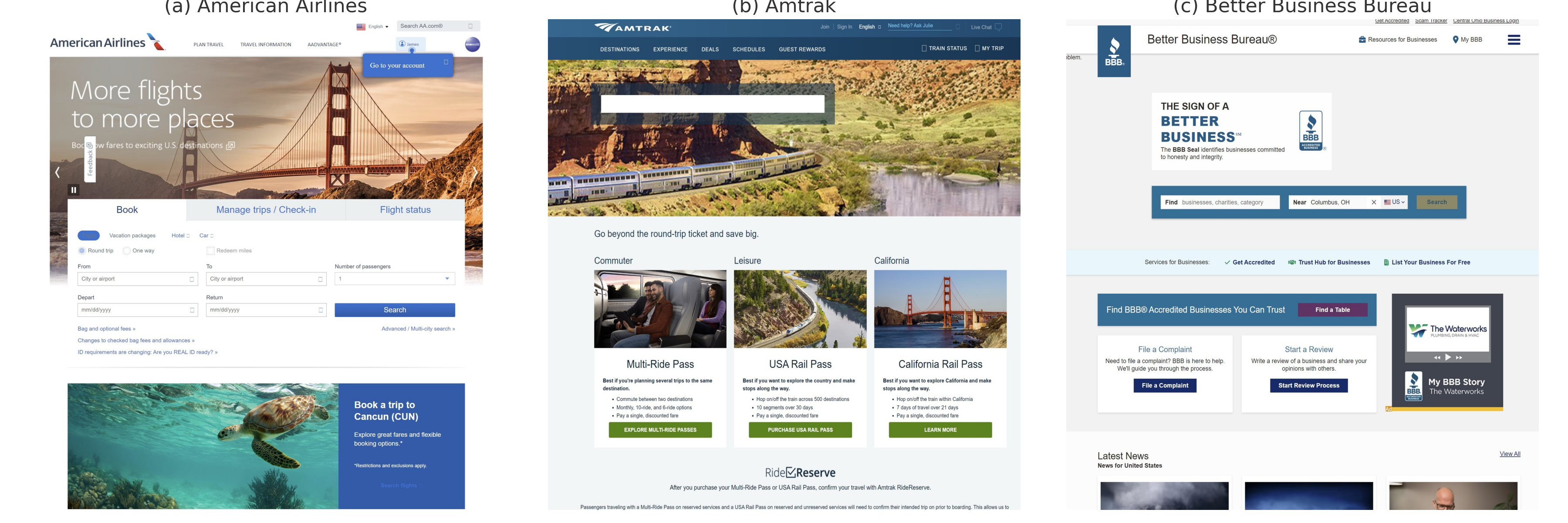}
  \caption{Illustrative examples from the MIND2WEB dataset showing the diversity in tasks and domains.}
  \label{fig:dataset_examples}
\end{figure*}

\section{Benchmark Data Set}
We discuss our benchmark data curation in section \ref{sec:benchmark-curation}, The list of data samples is included in our supplementary code. After extracting the code, you'll find a single CSV file in the 'data' folder, listing all 1,000 selected samples, including those with duplicate elements where one is the ground truth. In the 'supplement' folder, you'll find two result dataframes: 'grounding\_failures\_no\_duplicate.csv,' which includes failed samples without duplicates, and 'grounding\_no\_duplicates.csv,' which contains all 777 samples without duplicates.

\section{Benchmark Data Curation}
\label{sec:benchmark-curation}
We analyzed 8,628 samples from the Mind2Web Test dataset, out of which our DOM SUIU failed to execute on 487 test cases, representing approximately 5.64\% of the dataset. Root cause analysis revealed that these failures were due to corrupt MHTML files and missing ``node-buckeye-id'' annotations with corresponding ``action\_uid'' annotations. We focused only on positive candidates, excluding negative ones, to validate that our SUIU algorithm correctly assigns bounding box values consistent with the calculations performed during the Mind2Web dataset creation. This mapping of our SUIU candidate elements to Mind2Web's identified positive elements is a crucial step in establishing the SUIU ground truth for our end-to-end experiment.

To ensure accuracy, we computed the bounding box Jaccard Index along with additional features to confidently identify a SUIU element as a ground truth positive candidate. In the course of this analysis, we further excluded 1,779 test cases: 956 samples were dropped because their comparison features fell below the required threshold, with further analysis indicating that these failures were due to significant horizontal or vertical skew. This discrepancy appears to stem from differences in rendering between our desktop environment and the MHTML sampling done during Mind2Web dataset generation. Additionally, 823 samples were excluded because the ground truth positive candidate specified by Mind2Web was not present among the SUIU-extracted candidate elements.

Ultimately, we were left with 6,362 valid test cases for our experiment, where each set of candidate elements from our SUIU included the target ground truth element marked by a dedicated attribute.

From these verified test cases, we sampled a 15\% subset, resulting in 995 meaningful samples that maintained a distribution comparable across all test splits.

\begin{figure}[ht!]
    \centering
    \includegraphics[width=\linewidth]{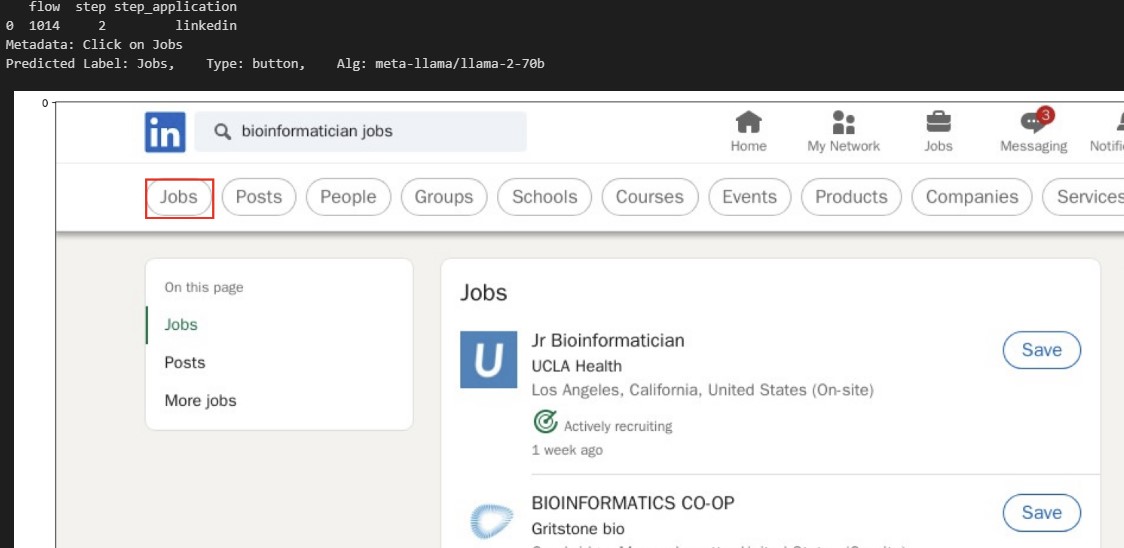}
    \caption{Duplicate: ``Click on Jobs''}
    \label{fig:suiu-duplicates}
\end{figure}

During our grounding evaluation, we identified 218 samples—approximately 21\% of the total—that highlight a significant challenge in planning tasks: the presence of duplicates. A sample is considered a duplicate if the ground truth element has one or more identical additional elements that can cause current solutions to fail without any additional external knowledge. In Figure \ref{fig:suiu-duplicates}, for example, the application contains duplicate clickable elements—a button and a link, both labeled ``Jobs.'' This duplication poses a challenge for both grounding and planning, as selecting between the two becomes a matter of chance, with success rates depending on a coin flip. We plan to address this challenge in future work.

It's worth noting that our analysis also revealed that the Mind2Web Test and Train splits each contain a similar percentage of duplicates, both around 21\%.

\section{Grounding Error Analysis}
\label{sec:grd-err-anls}

We demonstrated that out of 777 samples, 703 were successful, leaving 74 samples that failed. We thoroughly analyzed these 74 failures to identify the root causes and categorized the errors into two main groups: 35\% of the errors were due to gaps in our own algorithm, while the remaining 65\% were attributed to inherent issues within the Mind2Web dataset.

The algorithmic gaps we identified include some relatively simple problems that can be addressed with minor adjustments, while others require more complex or in-depth solutions. A detailed discussion of these gaps will be addressed in other venues.

The inherent issues within the Mind2Web dataset were further classified into several core problems, including clickable objects, ambiguous target elements, nested elements (``box in a box'') and offline flow equal trajectories to success. Each of these issues is discussed in its own subsection.

\begin{figure}[ht!]
    \centering
    \begin{subfigure}{\linewidth}
        \includegraphics[width=\linewidth]{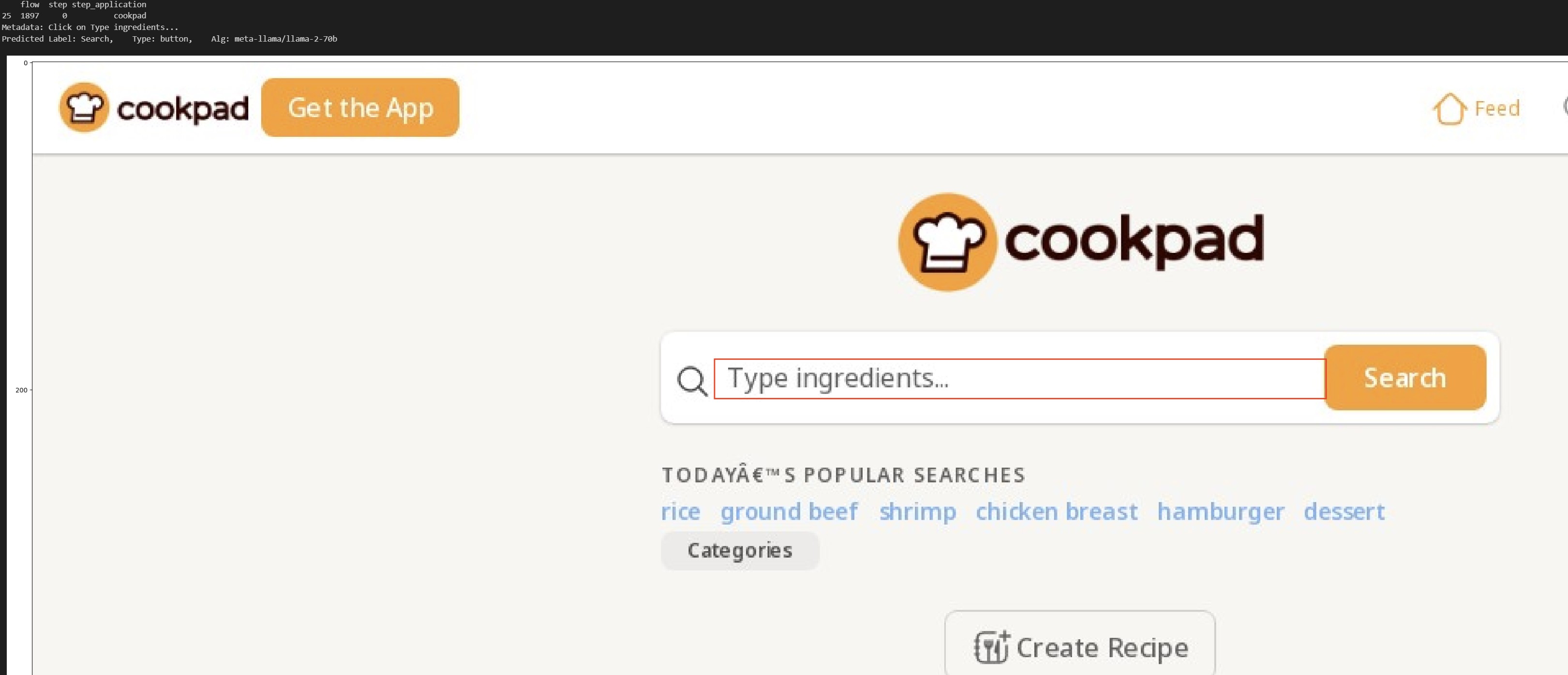}
        \caption{Instruction: ``Click on Type Ingredients...''}
        \label{fig:Click-on-typeable-object}
    \end{subfigure}
    \begin{subfigure}{\linewidth}
        \includegraphics[width=\linewidth]{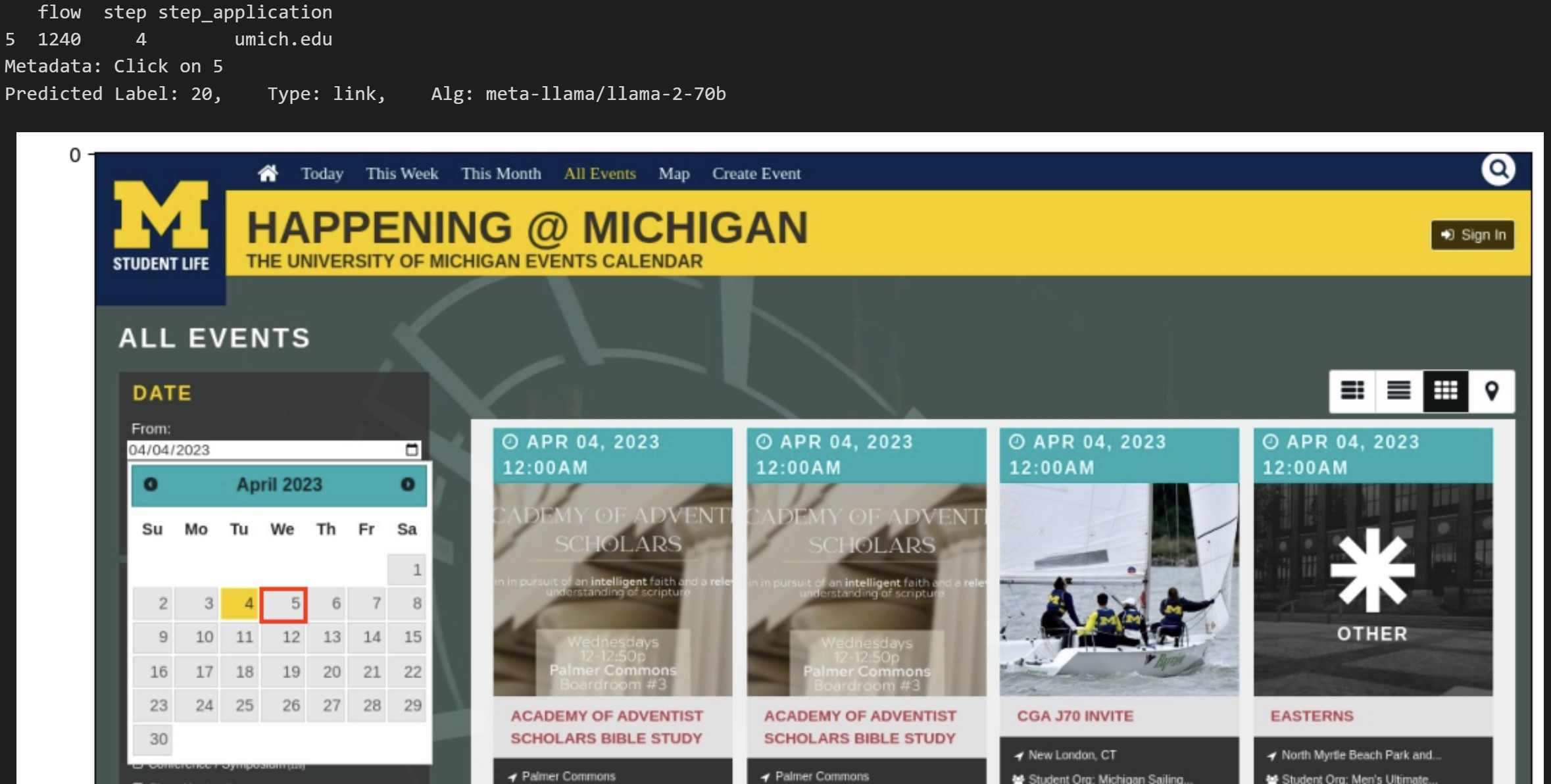}
        \caption{Instruction: ``Click on 5''}
        \label{fig:Ambigous-target-element}
    \end{subfigure}
    \begin{subfigure}{\linewidth}
        \includegraphics[width=\linewidth]{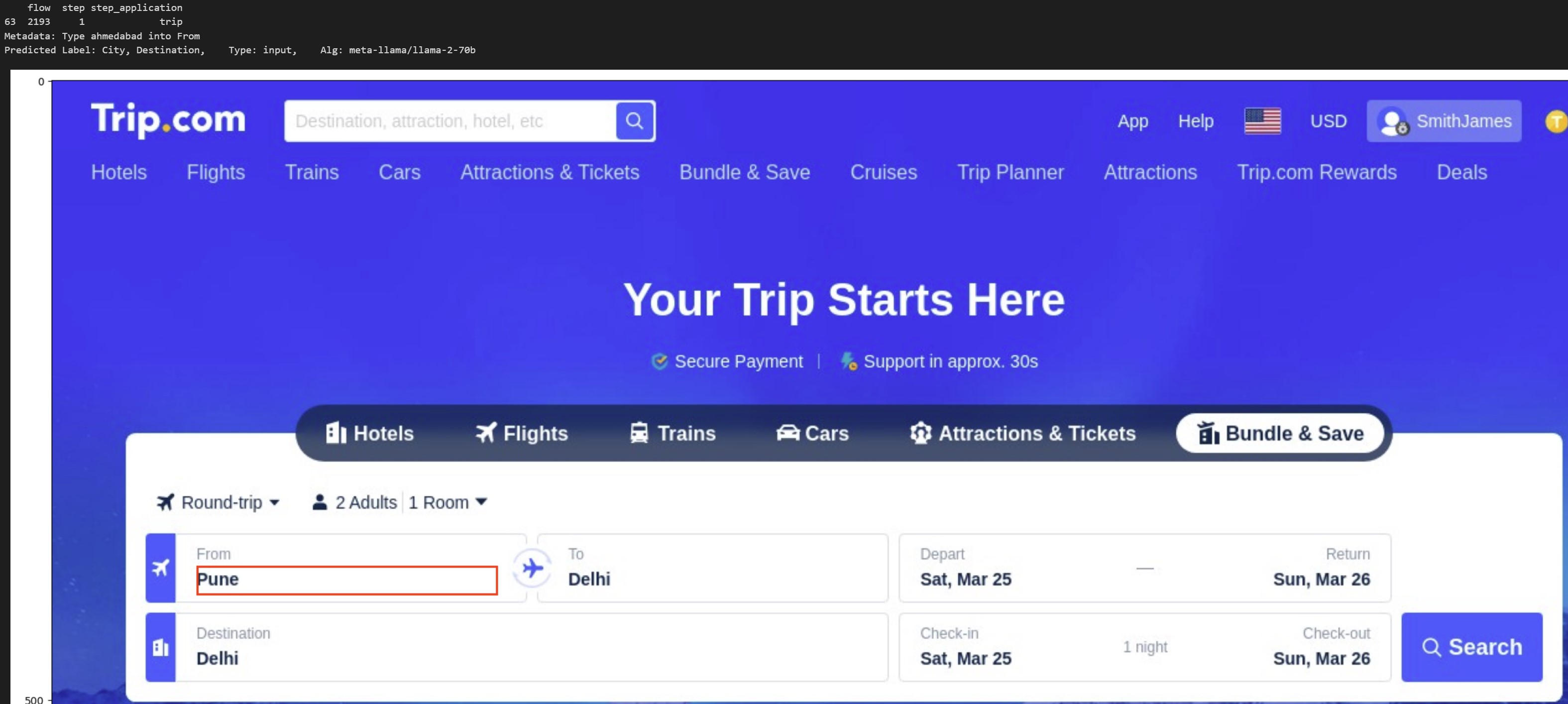}
        \caption{Eqaul Flow Trajectory}
        \label{fig:equal-trajectory-to-success}
    \end{subfigure}
    \caption{Examples of element grounding errors: (a) Click on a typeable object, (b) Ambiguous target element, (c) Equal Flow Trajectory.}
    \label{fig:composite_1}
    \end{figure}

    \begin{figure}[ht!]
    \centering
    \begin{subfigure}{\linewidth}
        \includegraphics[width=\linewidth]{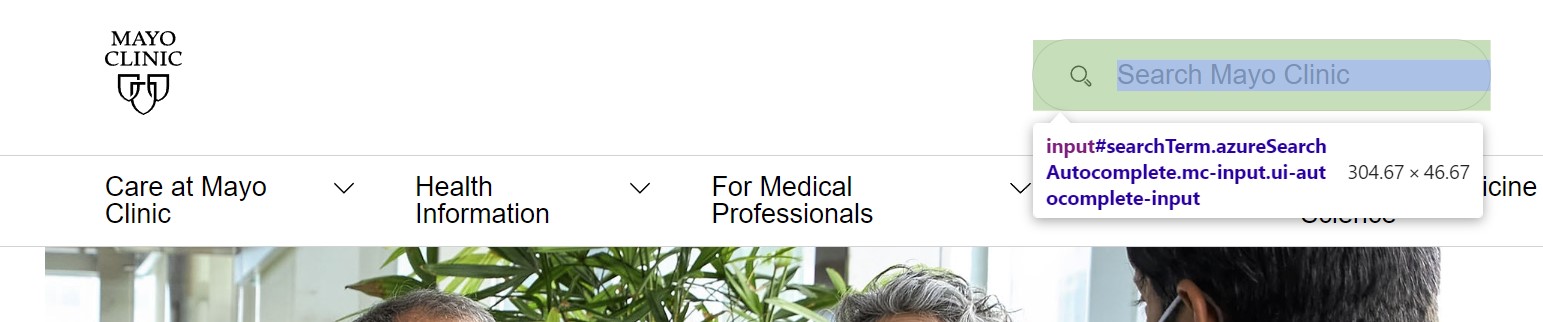}
        \caption{Input field marked as Ground Truth}
        \label{fig:box-in-box-inspect}
    \end{subfigure}
    \begin{subfigure}{\linewidth}
        \includegraphics[width=\linewidth]{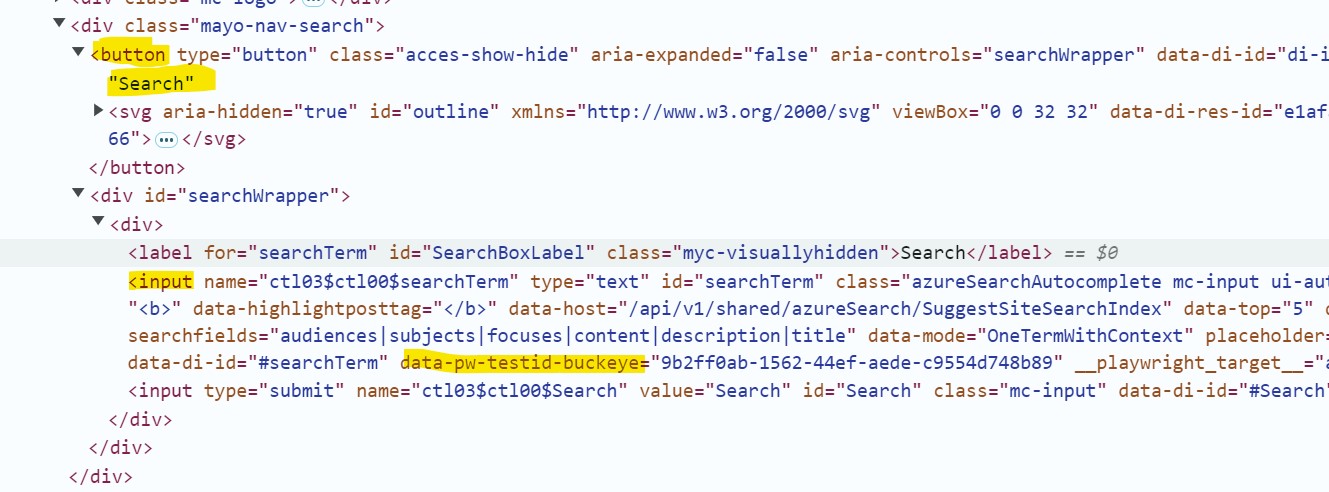}
        \caption{Button inside the box}
        \label{fig:box-in-box-html}
    \end{subfigure}
    \begin{subfigure}{\linewidth}
        \includegraphics[width=\linewidth]{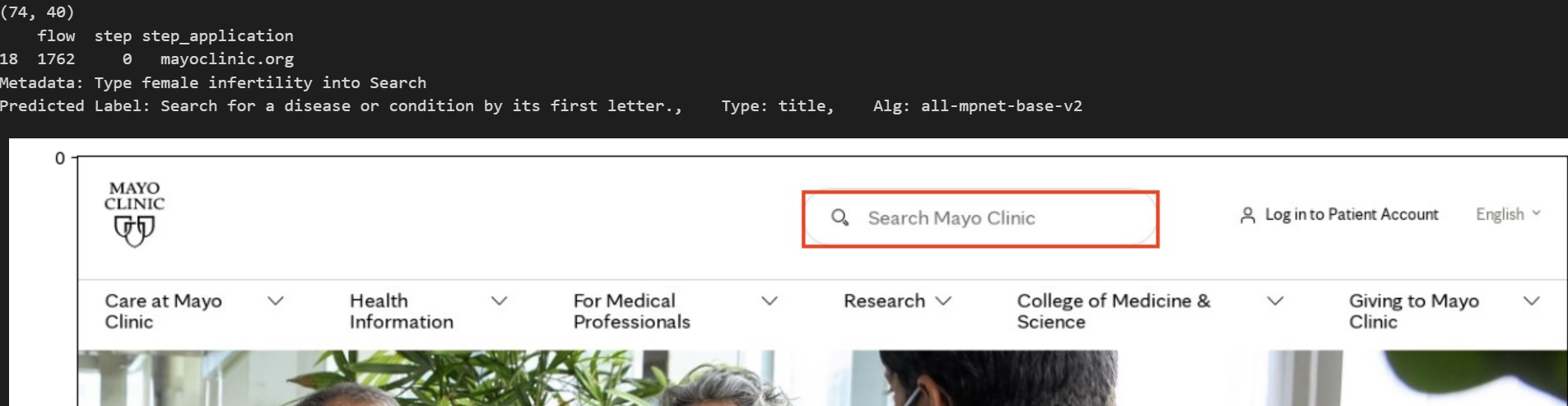}
        \caption{Instruction: ``Click on Search''}
        \label{fig:box-in-box}
    \end{subfigure}    
    \caption{Box in a Box grounding error example shows (a) ground truth element is the input element (b) button element exists within the requested bounding box (c) Mind2Web GT bounding box.}
    \label{fig:composite_2}
    
\end{figure}

\subsubsection{Click on typeable object}
The Mind2Web annotator describes the action as a ``click'' operation on an element that is a typeable object. Figure \ref{fig:Click-on-typeable-object} provides an example where the referenced element is correctly described as ``Type Ingredients...''. However, the action ``Click'' typically applies to clickable elements like buttons or links, whereas the actual element in question is an input field that requires a ``Type'' action. This discrepancy leads to failures in correctly identifying the action, making it one of the more prevalent issues among the failed samples.


\subsubsection{Ambiguous target element}
Another category of dataset-related failures involves ambiguous descriptions of target elements, which make it difficult to identify the correct interactable elements. In Figure \ref{fig:Ambigous-target-element}, for instance, the instruction ``Click on 5'' is unclear. A more specific instruction, such as ``Click on 5 beds'' or ``Click on July 5th'', would be needed to avoid confusion.


\subsubsection{Box in a Box}
Another category of failures we identified can be described as the ``box-in-box'' issue. 
This occurs when the annotator selects elements on the screen that correspond to a 
bounding box in the rendered HTML containing multiple nested elements. While the 
annotator marks one of these elements as the ground truth, the bounding box often 
encompasses several interactive elements, leading to potential ambiguity.

Figure \ref{fig:box-in-box} illustrates this issue. The red outline in the figure represents the 
bounding box, which contains multiple HTML elements. As shown in Figure \ref{fig:box-in-box-inspect}, 
the ground truth element is an input element. However, the same bounding box also includes 
a button element, as evidenced in the HTML inspect snippet in Figure \ref{fig:box-in-box-html}. 
Interestingly, in this case, the button element may actually be a more appropriate 
representation of the action ``Click on Search'' described in the step. This discrepancy 
highlights the complexity of accurately mapping user instructions to specific HTML 
elements, especially in cases where multiple interactive elements are in close proximity.




\subsubsection{Equal Flow Trajectory Success}
A critical category that influences the measurement of flow success rates is tied to the nature of offline experiments. In these experiments, we are restricted to a specific trajectory for success, whereas, in dynamic planning, multiple valid trajectories can lead to the same outcome.

For example, in Figure \ref{fig:equal-trajectory-to-success} we observe a typical flight reservation app where a customer can choose to fill in the ``From'' field before the ``Destination'' field, or vice versa. The planner may select any order, but the action history provides no useful guidance on which trajectory to follow. This restriction often leads to failures during immediate offline grounding.



\section{Partial Knowledge based on Pre-Planning} \label{appendix:preplanning}
\subsection{Prompt for the Pre-Planning LLM}
\textit{``Given a description, {flow}, of a task, write a short paragraph that explains in general how to solve the task, assuming you are already on the desired web page. Avoid suggesting the use of a search bar, searching, or scrolling.''}

\subsection{Examples or Pre-Planning output}

\subsubsection{A}
\textbf{Flow:} Find a flight for three adults from New York to New Orleans on April 14, return on April 16.
\textbf{General instruction:} To find a flight for three adults from New York to New Orleans on April 14, returning on April 16, navigate directly to the booking section of the webpage and enter the necessary details. First, select the option for round-trip. Next, under the ``From'' field, choose or type ``New York,'' and for the ``To'' field, select or type ``New Orleans.'' Then, input the departure date as April 14 and the return date as April 16. Ensure that the number of passengers is set to three adults. After entering all these details, proceed to review the available flight options by clicking the submission button, which will display a list of flights that match the criteria.

\subsubsection{B}
\textbf{Flow:} Find a highest rated dealer for Cadillac with rating above 4 star within 20 miles of zip 60606.
\textbf{General instruction:} First, navigate to the section of the webpage where dealer information is listed. Look for the filters or options to refine your search criteria. Set the filters to display only Cadillac dealers. Next, adjust the distance or location settings to limit the results to a radius of 20 miles from the zip code 60606. After this, locate and apply the filter for dealer ratings, ensuring you select only those with ratings above 4 stars. Finally, review the filtered results and identify the highest-rated dealer based on the provided ratings.

\subsubsection{C}
\textbf{Flow:} Signup for virtual healthcare visit.
\textbf{General instruction:} To signup for a virtual healthcare visit, start by carefully reading through the webpage to look for any obvious links or buttons related to booking appointments or virtual visits. Once located, click on the relevant link or button to initiate the sign-up process. You'll likely encounter a form requiring personal information, such as name, contact details, and possibly medical history or insurance information. Fill out all the required fields accurately. After providing the necessary information, submit the form and wait for confirmation. Ensure you set a convenient date and time for your virtual visit and note any specific instructions or preparations needed for the appointment.

\subsubsection{D}
\textbf{Flow:} Search for a comment about hip hop on the post about meme with the most comments from the last week
\textbf{General instruction:} To find a comment about hip hop on a post about a meme with the most comments from the last week, start by identifying the post in question. Since the goal is to locate a specific comment from a large pool, navigate directly to the comments section, which is typically located below the main content of the post. Once in the comments section, look for any options that allow you to filter or sort the comments. Choose a sorting method such as ``most recent'' or ``most liked'' to help manage the flow of comments. Read through the comments systematically, paying attention to any discussion or mentions of hip hop. Comments related to hip hop may mention artists, lyrics, songs, or general opinions about the genre. Continue this methodical reading and filtering until you locate the relevant comment about hip hop.

\end{document}